\documentclass{article}

\IfFileExists{neurips_2026.sty}{
  \usepackage[preprint]{neurips_2026}
}{
  \IfFileExists{neurips_2025.sty}{
    \usepackage[preprint]{neurips_2025}
  }{
    \IfFileExists{neurips_2024.sty}{
      \usepackage[preprint]{neurips_2024}
    }{
      \IfFileExists{neurips_2023.sty}{
        \usepackage[preprint]{neurips_2023}
      }{
        \usepackage[margin=1in]{geometry}
      }
    }
  }
}

\usepackage{times}
\usepackage{latexsym}
\usepackage[T1]{fontenc}
\usepackage[utf8]{inputenc}
\usepackage{microtype}
\usepackage{inconsolata}
\usepackage{booktabs}
\usepackage{graphicx}
\usepackage{multirow}
\usepackage{siunitx}
\usepackage{hyperref}
\usepackage{enumitem}
\usepackage{amsmath}
\usepackage{amssymb}
\usepackage{placeins}
\usepackage{float}
\usepackage{dblfloatfix}

\title{Democratizing AI with Small Language Models:\
Structured Benchmarking and Parameter-Efficient Fine-Tuning for Local Deployment}

\author{Daniel Cersosimo \\
  Binghamton University \\
  \texttt{dcersos1@binghamton.edu}}

\begin{document}
\maketitle

\begin{abstract}
AI democratization is not primarily a question of matching frontier-scale generality; it is a question of whether capable models can be selected, audited, and specialized under hardware and governance constraints that ordinary institutions can actually satisfy. This paper studies that problem through a controlled evaluation of nine open-weight language models between 135M and 3B parameters on a 1,085-example, 16-topic multiple-choice benchmark designed for structured local deployment. The benchmark emphasizes symbolic precision, constrained formatting, extraction, and short-horizon semantic decision making under a strict one-letter output protocol. A shared parameter-efficient fine-tuning pipeline then adapts a subset of models using 4-bit NF4 quantization with DoRA/LoRA-style adapters on an NVIDIA L4-class budget. In base evaluation, Qwen Coder 3B leads at 75.67\% strict accuracy, followed by Qwen2.5 1.5B at 67.10\%, Qwen3.5 2B at 64.98\%, and Granite 3.3 2B at 64.61\%. On the shared 108-example held-out fine-tuning split, adaptation improves Qwen Coder 3B by +26.85 points, SmolLM2 1.7B by +25.92, Qwen2.5 1.5B by +19.44, SmolLM2 360M by +10.18, and SmolLM2 135M by +5.55. Across ranking, topic-level heterogeneity, difficulty strata, failure composition, efficiency frontiers, and topic-conditioned transfer, the same conclusion recurs: a disciplined workflow of benchmark construction, cross-model evaluation, and low-cost specialization already makes a subset of sub-3B models viable as local experts for structured niche workloads.
\end{abstract}

\section{Introduction}
The contemporary discussion of AI democratization is often framed as a question of access to increasingly large frontier models. For practical deployment, however, the more consequential question is narrower and more operational: can useful systems be built, evaluated, and governed by actors who do not possess hyperscale infrastructure? Prior work on compute concentration, environmental cost, and foundation-model governance shows why this question matters both technically and socially [21, 22, 23, 24]. If model quality is only attainable under centralized training and serving regimes, participation remains structurally limited. If strong task performance can instead be achieved with compact open models that fit on accessible hardware and can be adapted locally, then ownership, privacy, and experimentation broaden substantially.

This paper studies that latter possibility. The focus is not unrestricted open-ended dialogue, but structured deployment: settings in which a model must return a short, schema-constrained answer and where errors are costly because downstream systems assume a valid output. Examples include lightweight extraction, label assignment, bounded decision support, and rule-sensitive transformations. These are precisely the kinds of workloads for which small language models (SLMs) are attractive: they are inexpensive to run, can often be hosted locally, and can be specialized without full-model retraining.

The analysis proceeds in two stages. First, nine open-weight models between 135M and 3B parameters are evaluated on a controlled 1,085-example benchmark spanning 16 multiple-choice task families. The benchmark was intentionally assembled to mix symbolic tasks, semantic categorization, scenario-based decisions, and formatting-sensitive procedures under a common prompt and scoring interface. Second, a shared parameter-efficient fine-tuning (PEFT) pipeline adapts a subset of these models using 4-bit quantization and DoRA/LoRA-style adapters, allowing direct measurement of how much local specialization can change viability for a narrowly defined structured workload.

The central claim is that democratized AI should be understood as a reproducible workflow rather than as a single model-selection event. The workflow is: define a bounded use case, construct an evaluation set aligned to that use case, compare open SLMs under a fixed protocol, fine-tune promising candidates with low-cost adapters, and then deploy the resulting model as a local expert only where its audited competence is adequate. The contribution of this paper is therefore both empirical and methodological.

\paragraph{Contributions.}
\begin{itemize}[leftmargin=*]
\item A controlled 16-topic benchmark for structured MCQ evaluation that stresses symbolic fidelity, schema compliance, and compact semantic reasoning under a shared output contract.
\item A comparative empirical study of nine open-weight models from 135M to 3B parameters using strict accuracy, topic-conditioned breakdowns, difficulty structure, failure composition, and deployment-oriented efficiency analysis.
\item A script-faithful PEFT study using quantized DoRA/LoRA-style adaptation on an NVIDIA L4-class budget, with shared split boundaries and deterministic post-tuning evaluation.
\item A deployment-oriented interpretation of the results that frames efficient SLMs as local experts and PEFT as a practical mechanism for converting general compact models into niche task specialists.
\end{itemize}

\section{Related Work}
This paper lies at the intersection of parameter-efficient adaptation, open small-model ecosystems, and evaluation methodology for constrained deployment.

\paragraph{Parameter-efficient adaptation.}
Low-rank adaptation made post-training accessible by restricting optimization to a low-dimensional update subspace [1]. QLoRA subsequently demonstrated that 4-bit quantization can preserve adaptation quality while lowering memory enough to make fine-tuning practical on modest hardware [2]. DoRA refines this line by decoupling directional and magnitude updates, improving stability in low-rank settings [3]. The pipeline used here inherits directly from this literature: frozen quantized backbones, trainable low-rank adapters, and completion-only supervision targeted at the answer span.

\paragraph{Open small-model ecosystems.}
Compact open model families have improved rapidly. SmolLM2 emphasizes data-centric training at multiple sub-2B scales [4]. OLMo2 foregrounds full-stack openness [5]. Qwen2- and Qwen2.5-family models show that aggressive pretraining and instruction alignment can produce highly competitive compact checkpoints [6, 7]. Parallel work on TinyLlama, Phi-3, and MobileLLM reinforces the broader trajectory: the performance frontier for small models is moving quickly enough that carefully selected sub-3B models are now plausible deployment candidates for bounded tasks [26, 27, 28].

\paragraph{Evaluation for democratized deployment.}
For local deployment, single-number benchmark reporting is insufficient. Formatting reliability, latency, token cost, and error concentration matter because the intended use case is not generic conversation but structured integration into low-budget systems. This paper therefore treats strict correctness, difficulty-conditioned performance, failure composition, and efficiency frontiers as primary evaluation objects rather than as supplementary diagnostics.

\section{Benchmark Construction}
\subsection{Controlled Benchmark Design}
The benchmark contains 1,085 rows with fields for task identifier, topic, question text, prompt scaffold, and gold answer. Every item terminates in a constrained output slot instructing the model to emit exactly one option letter and nothing else. This design is deliberate. In structured deployment, formatting is not separable from correctness; a semantically recoverable but malformed response is often operationally unusable. The benchmark therefore encodes format validity into the scoring rule rather than treating it as a secondary quality axis.

Benchmark generation follows a category-first process. Topic families were selected to cover symbolic precision, extraction, constrained formatting, rule execution, lightweight classification, and short-context decision making. A scripted generator then instantiated diverse item variants within each family while preserving a constant prompt interface. The resulting dataset contains 158 unique stems expressed through a fixed set of instruction and option templates. The benchmark is thus intentionally controlled: it is broad enough to produce meaningful cross-model separation, but regular enough that a low-resource lab can reproduce results and interpret failure modes without confounding prompt-schema drift.

This design choice is also the principal scope condition on the empirical claims. The dataset is not presented as an unconstrained natural-language understanding benchmark. It is a controlled-structure instrument for model selection under a shared interface. Consequently, the reported conclusions are strongest for structured MCQ-style deployment and should be read as dataset-conditional rather than universal claims about model superiority.

\subsection{Topic Inventory and Label Structure}
The benchmark spans 16 topics and mixes binary, ternary, and four-way choice spaces. Table~\ref{tab:topic_design} summarizes the topic inventory, choice configuration, tested capability, and representative task stems.

\begin{table*}[t]
\centering
\scriptsize
\begin{tabular}{p{0.14\linewidth} r c p{0.24\linewidth} p{0.25\linewidth}}
\toprule
\textbf{Topic} & \textbf{N} & \textbf{Choices} & \textbf{Capability Tested} & \textbf{Representative Task Stem} \\
\midrule
Rule Application & 110 & A/B/C/D & Deterministic rule execution under fixed label options & ``Classify numbers as even or odd.'' \\
Extraction & 100 & A/B/C/D & Exact retrieval of structured signals from text & ``Extract all email addresses.'' \\
Scenario Decision & 97 & A/B & Short-horizon contextual action choice & ``Car wash drive or walk.'' \\
Counting & 90 & A/B/C/D & Character/token-level symbolic precision & ``How many times does the letter r appear in strawberry?'' \\
Constraint Following & 85 & A/B & Adherence to explicit output/format constraints & ``Return exactly one word.'' \\
Formatting & 80 & A/B/C/D & Canonicalization and transformation fidelity & ``Sort items alphabetically.'' \\
Topic Classification & 61 & A/B/C/D & Domain-level semantic categorization & ``Classify topic.'' \\
Sentiment Classification & 60 & A/B/C & Polarity interpretation under short context & ``Classify sentiment.'' \\
Risk Classification & 51 & A/B/C & Coarse risk-level categorization & ``Classify risk level.'' \\
NLI Classification & 51 & A/B/C & Relation inference (entail/neutral/contradict) & ``Classify relationship.'' \\
Label Selection & 50 & A/B & Schema-bound label mapping & ``Choose support status.'' \\
Binary Classification & 50 & A/B & Two-way semantic discrimination & ``Classify as urgent or not\_urgent.'' \\
Multiple Choice Reasoning & 50 & A/B/C & Comparative option-level reasoning & ``Choose best summary label.'' \\
Classification (contradiction/not) & 50 & A/B & Compact logical discrimination & ``Choose contradiction or not.'' \\
Mystery Reasoning & 50 & A/B/C & Lightweight narrative inference from sparse evidence & ``Choose the culprit.'' \\
Goal-Aware Choice & 50 & A/B & Utility-aligned action selection & ``Choose best action.'' \\
\bottomrule
\end{tabular}
\caption{Consolidated topic design matrix for the v3 benchmark: counts, choice-space configuration, tested capability, and representative task stems.}
\label{tab:topic_design}
\end{table*}

The label structure is intentionally non-uniform. There are 611 four-way items, 312 binary items, and 162 ternary items; global gold-label frequencies are A=413, B=353, C=200, and D=119. These properties are operationally important. First, the nominal chance rate is topic-dependent. Second, aggregate accuracy can be subtly influenced by answer prior bias if a model disproportionately emits high-frequency labels. The evaluation therefore emphasizes topic-resolved and difficulty-conditioned analysis rather than relying on a single corpus-wide rank.

\subsection{Representative Examples and Benchmark Scope}
Three examples illustrate the character of the benchmark. A counting item asks how many times the letter \texttt{r} appears in \texttt{strawberry}, probing symbolic fidelity under trivial world knowledge but non-trivial procedural exactness. An extraction item asks the model to select the option that exactly recovers all email addresses from a short passage, testing set-level precision rather than approximate paraphrase. A scenario-decision item asks whether a person should walk or drive to a nearby car wash under specified conditions, testing bounded context-sensitive action selection.

These examples are deliberately modest in surface form. The objective is not to maximize lexical novelty, but to expose whether a compact model can repeatedly satisfy a strict interface across heterogeneous micro-tasks. That is the relevant desideratum for many local deployments: not open-ended eloquence, but stable compliance with a narrow decision contract.

\section{Model Suite}
The evaluated model suite contains nine checkpoints spanning 135M to 3B parameters, chosen to cover several active open-model lineages and a meaningful range of training and alignment priors.

\paragraph{SmolLM2 family.}
The SmolLM2 135M, 360M, and 1.7B instruct checkpoints are particularly informative because they hold model family identity relatively constant while varying scale [4, 8, 9, 10]. Their reported training recipe is strongly data-centric, with very large token budgets relative to parameter count and post-training alignment stages including supervised fine-tuning and preference optimization. In this study, they serve as a direct test of how far careful training can push compact models on structured tasks.

\paragraph{Qwen family.}
Qwen2.5 1.5B-Instruct, Qwen2.5-Coder 3B-Instruct, and Qwen3.5 2B represent the strongest compact performers in the suite [6, 7, 12, 13, 14]. Their documented architecture stack includes features such as RoPE, SwiGLU, RMSNorm, and grouped-query attention variants. The coder-oriented 3B model is especially relevant to this benchmark because structured MCQ tasks reward compact exact outputs, disciplined formatting, and local symbolic precision---all properties plausibly reinforced by code-heavy pretraining mixtures.

\paragraph{Complementary baselines.}
StableLM Zephyr 3B, Granite 3.3 2B-Instruct, and OLMo2 1B-Instruct contribute architectural and alignment diversity [5, 11, 15, 16]. Zephyr provides a chat-oriented instruction baseline with different preference-optimization lineage; Granite contributes a strong enterprise-oriented compact instruct model; OLMo2 offers a fully open 1B-class reference with high format reliability. Together, these models provide a useful contrast to the SmolLM2 and Qwen families and help identify which behaviors are family-specific rather than purely scale-driven.

\section{Materials and Methods}
\subsection{Base Inference and Scoring}
Base evaluation uses a deterministic inference harness. For each question $q$ and model $m$, the pipeline constructs the model-specific prompt representation, generates one short completion, normalizes the output conservatively, and records correctness, formatting validity, completion status, generated token count, and latency. Decoding is intentionally simple because the use case is structured classification rather than creative generation; stochasticity would only confound cross-model comparisons.

Strict correctness is defined at the item level. Let $c_{m,q}\in\{0,1\}$ denote semantic correctness and $f_{m,q}\in\{0,1\}$ denote format validity. The scored indicator is
\begin{equation}
  a_{m,q} = \mathbb{1}[c_{m,q}=1 \wedge f_{m,q}=1],
\end{equation}
with model-level strict accuracy
\begin{equation}
  A_m = \frac{1}{N}\sum_{q=1}^{N} a_{m,q}.
\end{equation}
This scoring rule is intentionally harsh. It reflects the deployment assumption that a malformed output is not operationally interchangeable with a correct one-letter answer.

The analysis layer also computes per-topic accuracy $A_{m,t}$ and item difficulty. Difficulty is operationalized directly from the evaluation table as the fraction of models that fail an item:
\begin{equation}
  d(q) = 1 - \frac{1}{M}\sum_{m=1}^{M} a_{m,q},
\end{equation}
where $M=9$ models in the base benchmark. Thus $d(q)=1$ corresponds to all-model failure and $d(q)=0$ to universal success.

\subsection{Analytic Views}
The plotting and tabulation stage derives several complementary views from the same long-form evaluation table. Topic anisotropy is summarized through a model-by-topic heatmap. Difficulty structure is summarized through corpus-wide histograms and topic-conditioned boxplots over $d(q)$. Error composition partitions outputs into correct, wrong-but-formatted, format failure, and incomplete responses. Inter-model structure is summarized through Kendall-$\tau$ correlation over topic-profile rankings. Deployment trade-offs are summarized through radar profiles, Pareto frontiers in accuracy-cost space, and full response-cost distributions.

These views were chosen because they answer different deployment questions. Global ranking answers which models are broadly strongest. Topic anisotropy answers where a locally weaker model may still be viable. Difficulty stratification answers whether gains are concentrated on trivial items or on genuinely contested ones. Pareto and cost analyses answer whether accuracy improvements survive a cost-sensitive decision rule.

\subsection{Fine-Tuning Protocol and Compute Target}
Fine-tuning targets an NVIDIA L4-class profile [20], not as a limiting artifact but as a representative tier of democratized compute: accessible cloud sessions and prosumer-grade accelerator budgets rather than multi-node training clusters. A single topic-stratified 80/10/10 split with seed 42 is shared across all adapted models to preserve comparability.

Each training example is serialized as a chat-style interaction in which the user message is the full benchmark prompt and the assistant message is the gold answer letter. This makes the adaptation objective isomorphic to the deployment interface. Evaluation after fine-tuning uses the same deterministic decoding and the same strict scoring rule as base evaluation.

The adaptation stack combines bitsandbytes 4-bit NF4 quantization [2, 17], PEFT low-rank adapters [1, 18], DoRA-enabled updates [3], and the TRL SFT trainer [19]. The operational rationale is straightforward: freeze the expensive backbone, update only a small trainable subspace, and focus the loss almost entirely on the answer span.

\subsection{Implemented QDoRA Objective and Configuration}
The PEFT configuration enables both DoRA and RSLoRA flags (\texttt{use\_dora=True}, \texttt{use\_rslora=True}) with dropout 0.05, no trainable bias, and scaling
\begin{equation}
  \alpha = 2r,
\end{equation}
where $r\in\{8,16,32\}$ is selected by GPU-memory tier.

In standard low-rank form, a frozen weight matrix $W$ is adapted as
\begin{equation}
  W' = W + \Delta W, \qquad \Delta W = BA,
\end{equation}
with $A \in \mathbb{R}^{r\times d}$ and $B \in \mathbb{R}^{k\times r}$. The trainable parameter count therefore scales with $r(d+k)$ rather than $kd$, which is the principal reason the procedure remains feasible on L4-class hardware.

The completion-only collator searches for the \emph{last} occurrence of the assistant-response template within the tokenized prompt and masks everything before it. Let $x_{1:n}$ denote the token sequence, $m$ the template length, and $p^{\star}$ the start index of the final template occurrence. The target labels are
\begin{equation}
  y_i =
  \begin{cases}
    -100, & i \le p^{\star}+m,\\
    x_i, & i > p^{\star}+m,
  \end{cases}
\end{equation}
with all labels masked if no template occurrence is found. Optimization then minimizes
\begin{equation}
  \mathcal{L} = -\frac{1}{|\Omega|}\sum_{i\in\Omega}\log p_{\theta}(x_i\mid x_{<i}),
  \qquad
  \Omega = \{i \mid y_i \neq -100\}.
\end{equation}
This objective is especially well matched to the benchmark because the desired completion is extremely short; the gradient budget is spent almost entirely on decision-boundary behavior rather than on reproducing prompt boilerplate.

Warmup is implemented in step units:
\begin{equation}
  \text{warmup\_steps}
  =
  \left\lfloor
  \rho \cdot
  \left\lfloor\frac{N_{\text{train}}}{B\cdot G}\right\rfloor
  \cdot E
  \right\rfloor,
\end{equation}
with warmup ratio $\rho=0.05$, epochs $E=3$, per-device batch size $B$, and gradient accumulation $G$.

\begin{table*}[t]
\centering
\small
\begin{tabular}{p{0.34\linewidth} p{0.58\linewidth}}
\toprule
\textbf{Component} & \textbf{Implemented Setting in \texttt{fine\_tune\_slms.py}} \\
\midrule
Data split & Topic-stratified 80/10/10 (train/val/test), seed 42 \\
Trainer & TRL SFTTrainer, eval/save each epoch, load best model at end by \texttt{eval\_loss} \\
Epochs / LR / Scheduler & 3 epochs, $2\times10^{-4}$, cosine schedule \\
Warmup / Regularization & warmup ratio 0.05 (step-based warmup formula above), weight decay 0.01, max grad norm 1.0 \\
Early stopping & patience = 3 evaluation rounds \\
Batching (tier-dependent) & T4: $B$=2, $G$=8; L4/A100-40GB: $B$=4, $G$=4; A100-80GB/H100: $B$=8, $G$=2 \\
Precision (tier-dependent) & T4: fp16; L4/A100-40GB: bf16 if supported else fp16; A100-80GB/H100: bf16 \\
Quantization & bitsandbytes 4-bit NF4, double quant enabled, compute dtype = tier dtype \\
Adapter mode & DoRA + RSLoRA enabled; LoRA dropout 0.05; bias none; $\alpha=2r$ \\
Adapter rank (tier-dependent) & $r=8$ (T4), $r=16$ (L4/A100-40GB default), $r=32$ (A100-80GB/H100) \\
Target modules & Auto-detected linear projection leaf names; in 4-bit path, bitsandbytes \texttt{Linear4bit} leaves excluding \texttt{lm\_head} \\
Memory control & Gradient checkpointing enabled via k-bit preparation path; \texttt{use\_cache=False} during training \\
Collator / supervision & Custom completion-only collator; loss applied only to answer span after last response-template occurrence \\
Decoding during evaluation & deterministic generation: \texttt{do\_sample=False}, \texttt{max\_new\_tokens=5}, \texttt{temperature=None}, \texttt{top\_p=None} \\
\bottomrule
\end{tabular}
\caption{Implemented fine-tuning configuration used in the sequential QDoRA loop. Values are reported exactly from the script-level configuration and control flow.}
\label{tab:ft_impl_cfg}
\end{table*}

\section{Base Results}
\subsection{Global Performance and Ranking Stability}
\begin{figure*}[!tbp]
\centering
\includegraphics[width=0.97\textwidth]{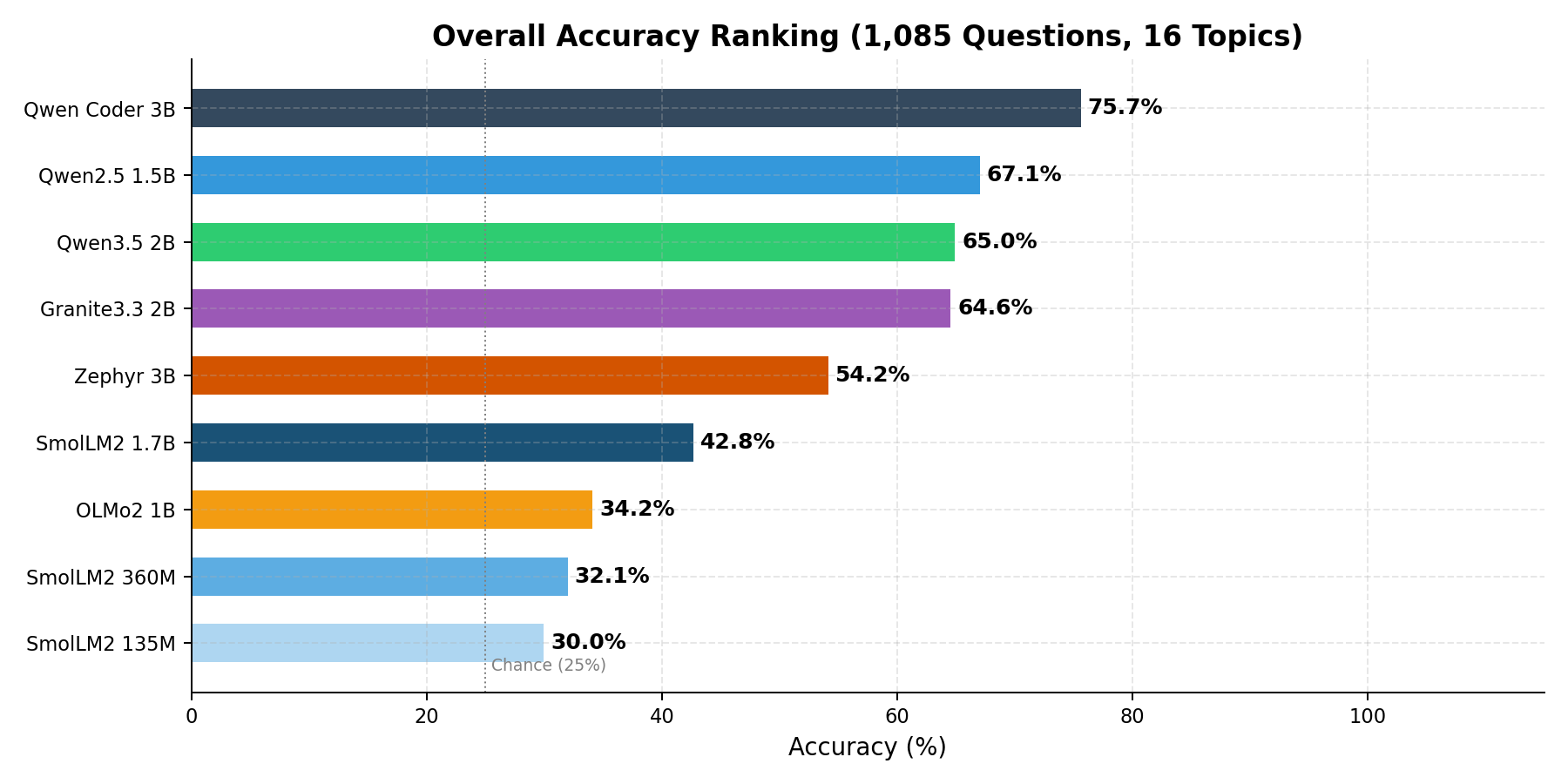}
\caption{Overall accuracy ranking. The separation between the leading Qwen-family models and the lower-capacity tier is explicit even under a strict output-validity scoring rule.}
\label{fig:leaderboard}
\end{figure*}

Table~\ref{tab:base} and Figure~\ref{fig:leaderboard} establish the coarse ranking. Qwen Coder 3B is the strongest base model at 75.67\% strict accuracy. Qwen2.5 1.5B, Qwen3.5 2B, and Granite 3.3 2B form the next tier at 67.10\%, 64.98\%, and 64.61\%, respectively. The small SmolLM2 variants and OLMo2 1B are materially weaker in aggregate semantics, but several of them already exhibit near-perfect format compliance and very low generation cost.

\begin{table*}[t]
\centering
\small
\begin{tabular}{lrrrrrr}
\toprule
Model & Accuracy (\%) & Format (\%) & Complete (\%) & Avg Tokens & Avg Latency (s) & Rank \\
\midrule
Qwen Coder 3B & 75.67 & 100.00 & 100.00 & 2.000 & 0.093 & 1 \\
Qwen2.5 1.5B & 67.10 & 100.00 & 100.00 & 2.000 & 0.072 & 2 \\
Qwen3.5 2B & 64.98 & 100.00 & 100.00 & 2.000 & 0.267 & 3 \\
Granite 3.3 2B & 64.61 & 98.53 & 100.00 & 2.039 & 0.099 & 4 \\
Zephyr 3B & 54.19 & 89.68 & 100.00 & 8.759 & 0.259 & 5 \\
SmolLM2 1.7B & 42.76 & 99.91 & 100.00 & 2.002 & 0.058 & 6 \\
OLMo2 1B & 34.19 & 99.26 & 99.91 & 1.999 & 0.053 & 7 \\
SmolLM2 360M & 32.07 & 99.91 & 100.00 & 2.008 & 0.074 & 8 \\
SmolLM2 135M & 30.05 & 91.98 & 100.00 & 2.249 & 0.079 & 9 \\
\bottomrule
\end{tabular}
\caption{Base benchmark summary on 1,085 examples. Strict accuracy requires both correct content and format-valid output.}
\label{tab:base}
\end{table*}

The more important conclusion is not merely that a 3B model wins, but that several sub-3B models already reach a practically relevant regime for structured workloads while retaining negligible output length and low latency. This is the key empirical precondition for democratization: there exists a compact model frontier that is meaningfully useful before fine-tuning begins.

Close gaps in the middle tier should nevertheless be interpreted cautiously. Exact McNemar testing on paired strict-correctness outcomes shows that Qwen3.5 2B (64.98\%) and Granite 3.3 2B (64.61\%) are not statistically distinguishable (discordant counts 203 vs. 199, $p=0.881$). Likewise, Qwen2.5 1.5B and Qwen3.5 2B are not significantly different under the same test (200 vs. 177, $p=0.257$). The top tier is therefore robustly separated from the small-model tail, but the internal ordering of some adjacent compact models should be read as approximate rather than absolute.

\subsection{Topic Anisotropy, Difficulty, and Failure Regime}
\begin{figure*}[!tbp]
\centering
\includegraphics[width=0.97\textwidth]{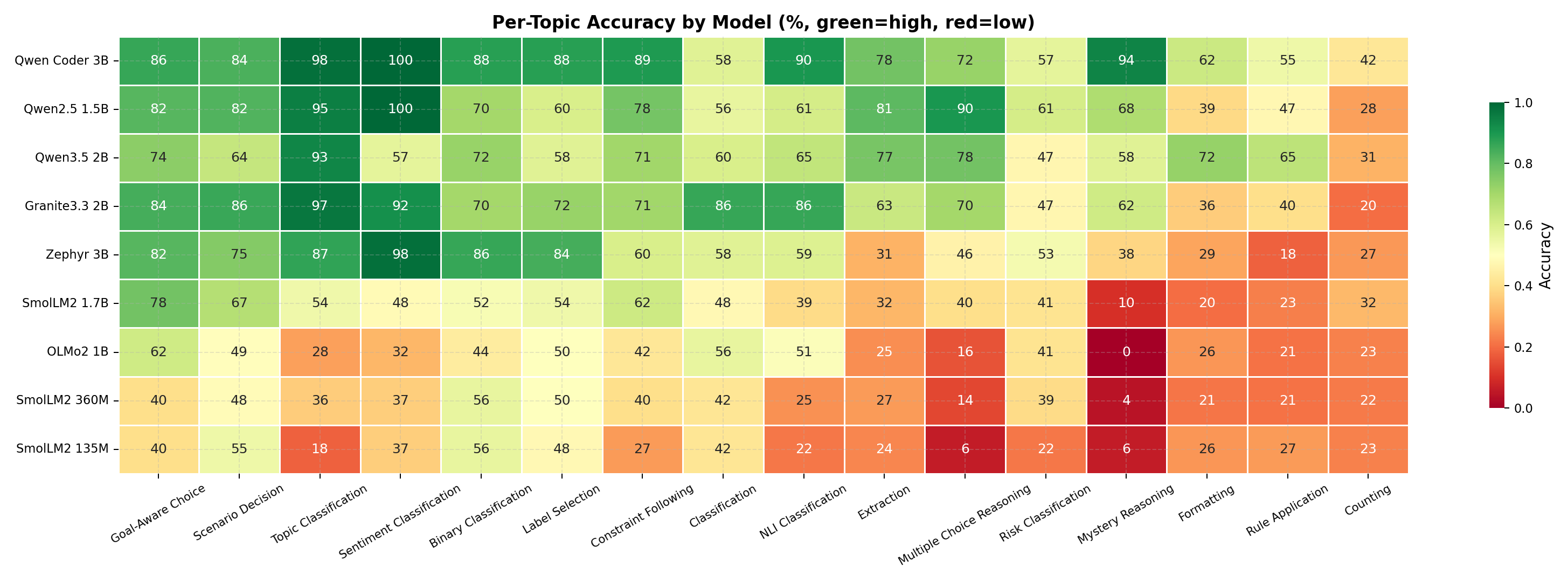}
\caption{Model-by-topic accuracy heatmap. The row-wise and column-wise structure makes capability anisotropy explicit: strong global models still have concentrated weaknesses, and lower-ranked models occasionally remain competitive on selected domains.}
\label{fig:topic_heatmap}
\end{figure*}

Aggregate ranking hides substantial structure. Figure~\ref{fig:topic_heatmap} shows that no model dominates uniformly across the 16-topic matrix. The top models maintain broad strength, but even they exhibit visible weakness on counting, formatting, and other structure-heavy categories. At the corpus level, the hardest topic is Counting (27.64\% average accuracy across models), followed by Rule Application and Formatting, whereas Goal-Aware Choice and Scenario Decision are materially easier.

This anisotropy matters operationally. A local deployment workflow should not treat model selection as a one-time leaderboard lookup. The more appropriate interpretation is conditional: a model is viable for a subset of task families, and the relevant decision is whether that subset aligns with the intended use case.

\begin{figure*}[!tbp]
\centering
\includegraphics[width=0.97\textwidth]{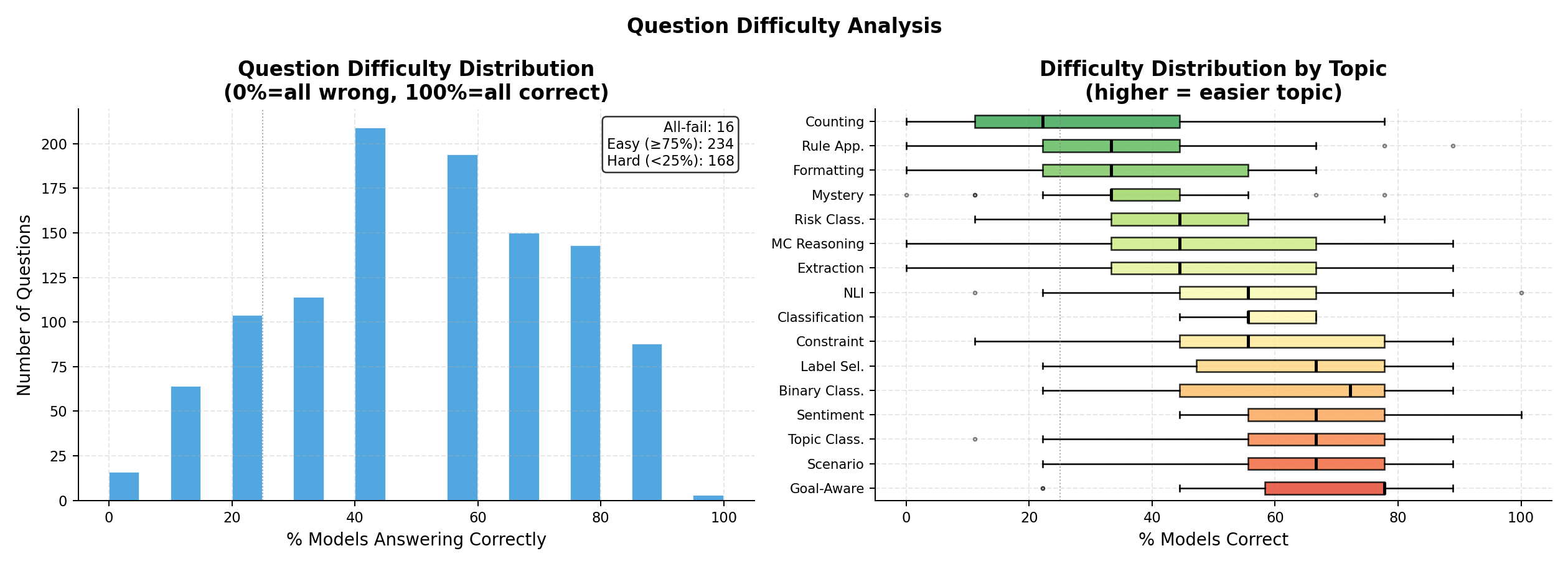}
\caption{Question difficulty distribution and topic-conditioned hardness. Difficulty is defined from the fraction of models that answer an item correctly, making the histogram and topic boxplots direct summaries of collective failure structure.}
\label{fig:difficulty}
\end{figure*}

Difficulty analysis sharpens the same point from the item side rather than the model side. Figure~\ref{fig:difficulty} summarizes the distribution of $d(q)=1-\text{pct\_correct}(q)$, where item difficulty is defined by the proportion of models that fail the same question. The benchmark contains a non-trivial hard tail, including items that nearly all models miss. This is crucial because improvements on universally easy items inflate global scores less meaningfully than improvements on contested items.

\begin{figure*}[!tbp]
\centering
\includegraphics[width=0.97\textwidth]{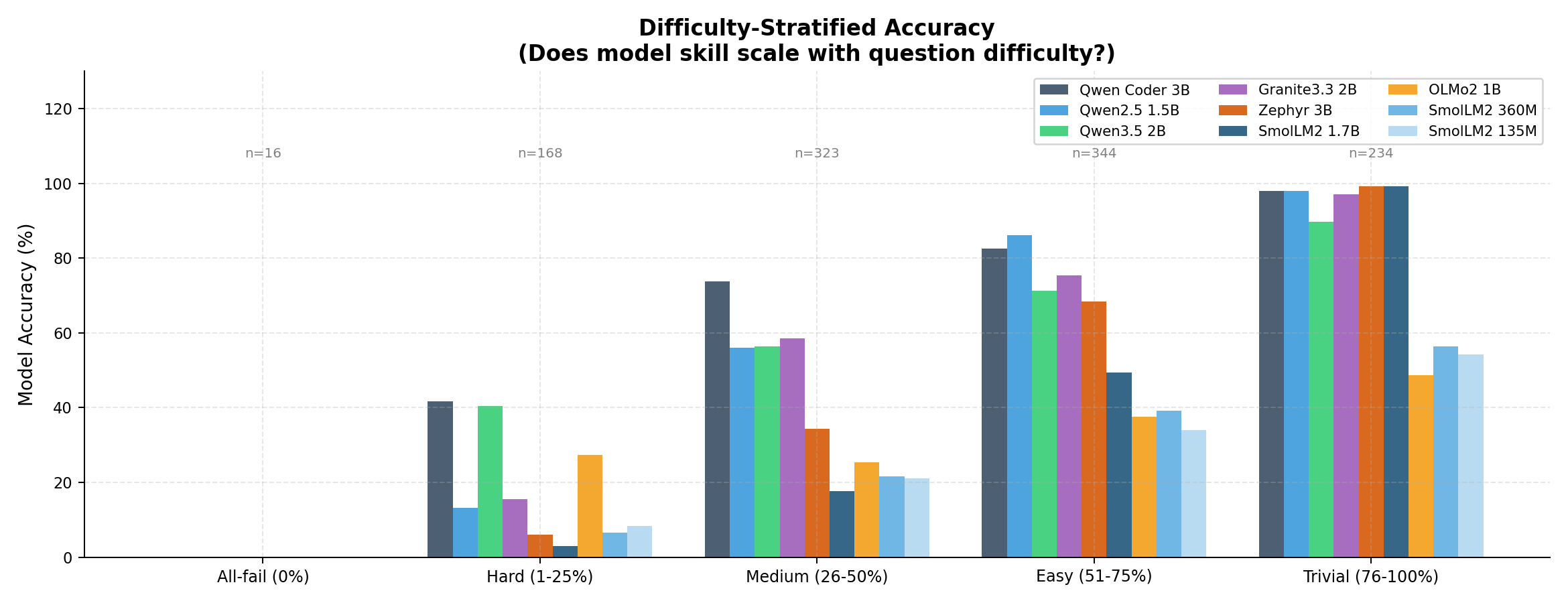}
\caption{Difficulty-stratified model accuracy. Stronger compact models preserve a smoother degradation profile as question difficulty increases, whereas lower-capacity models flatten quickly outside easy strata.}
\label{fig:difficulty_stratified}
\end{figure*}

Figure~\ref{fig:difficulty_stratified} then conditions each model's accuracy on difficulty buckets derived from the same quantity. Higher-capacity models retain separation precisely where it matters most: medium and hard strata rather than trivial items. This pattern is more informative than the global rank because it suggests that the best models are not simply exploiting answer priors or easy bins; they remain stronger under the parts of the benchmark that induce broad disagreement.

\begin{figure*}[!tbp]
\centering
\includegraphics[width=0.97\textwidth]{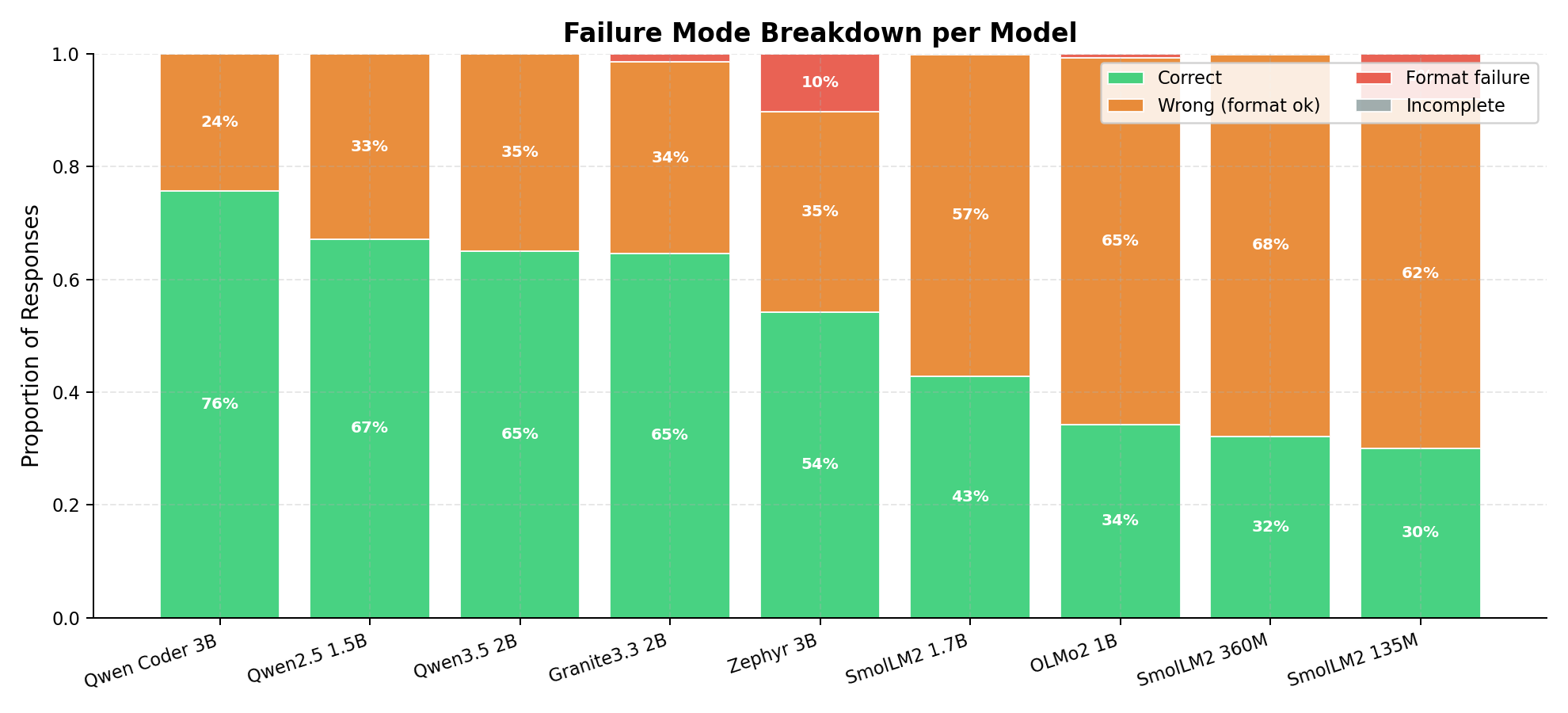}
\caption{Failure-mode composition by model. The prevailing error type is semantic misclassification under otherwise valid output formatting, not response omission.}
\label{fig:failure_modes}
\end{figure*}

Failure composition in Figure~\ref{fig:failure_modes} clarifies the dominant bottleneck. Across most models, the largest error mass is wrong-but-well-formatted output rather than missing output or gross format failure. That is a favorable regime for adaptation: the models already understand the interaction contract, so the remaining gap is primarily semantic discrimination rather than output-shape repair.

The combination of Figures~\ref{fig:topic_heatmap}--\ref{fig:failure_modes} suggests a precise interpretation of what compact models lack on this benchmark. The primary constraint is not the ability to emit a one-letter answer; it is the ability to preserve task-specific decision boundaries under symbolic and procedural variation. That distinction is central to the fine-tuning results that follow.

\subsection{Inter-Model Structure and Deployment Trade-offs}
\begin{figure}[!htbp]
\centering
\includegraphics[width=\linewidth]{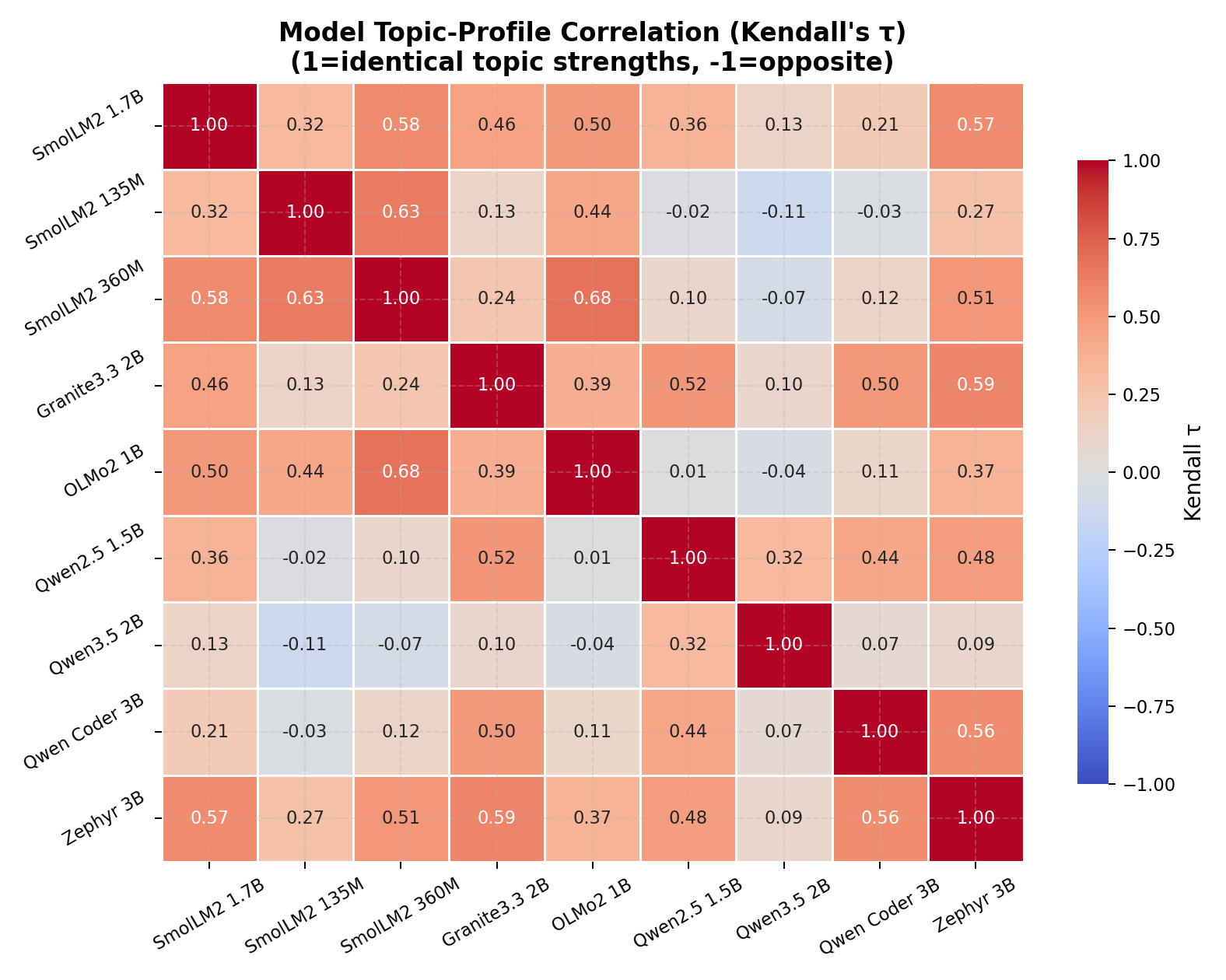}
\caption{Kendall-$\tau$ correlation over model topic profiles. Related model families cluster, but the matrix also reveals where topic-specialization patterns diverge enough to support complementary routing.}
\label{fig:topic_corr}
\end{figure}

Inter-model structure is not random. Figure~\ref{fig:topic_corr} reports Kendall-$\tau$ correlation between model topic profiles. Positive blocks indicate models that succeed and fail on similar topic subsets; weaker or negative values indicate complementary specialization. This is relevant for deployment because ensemble diversity should not be inferred from parameter count alone. Two models with similar overall accuracy can still be poor complements if their topic manifolds are too aligned.

\begin{figure}[!htbp]
\centering
\includegraphics[width=\linewidth]{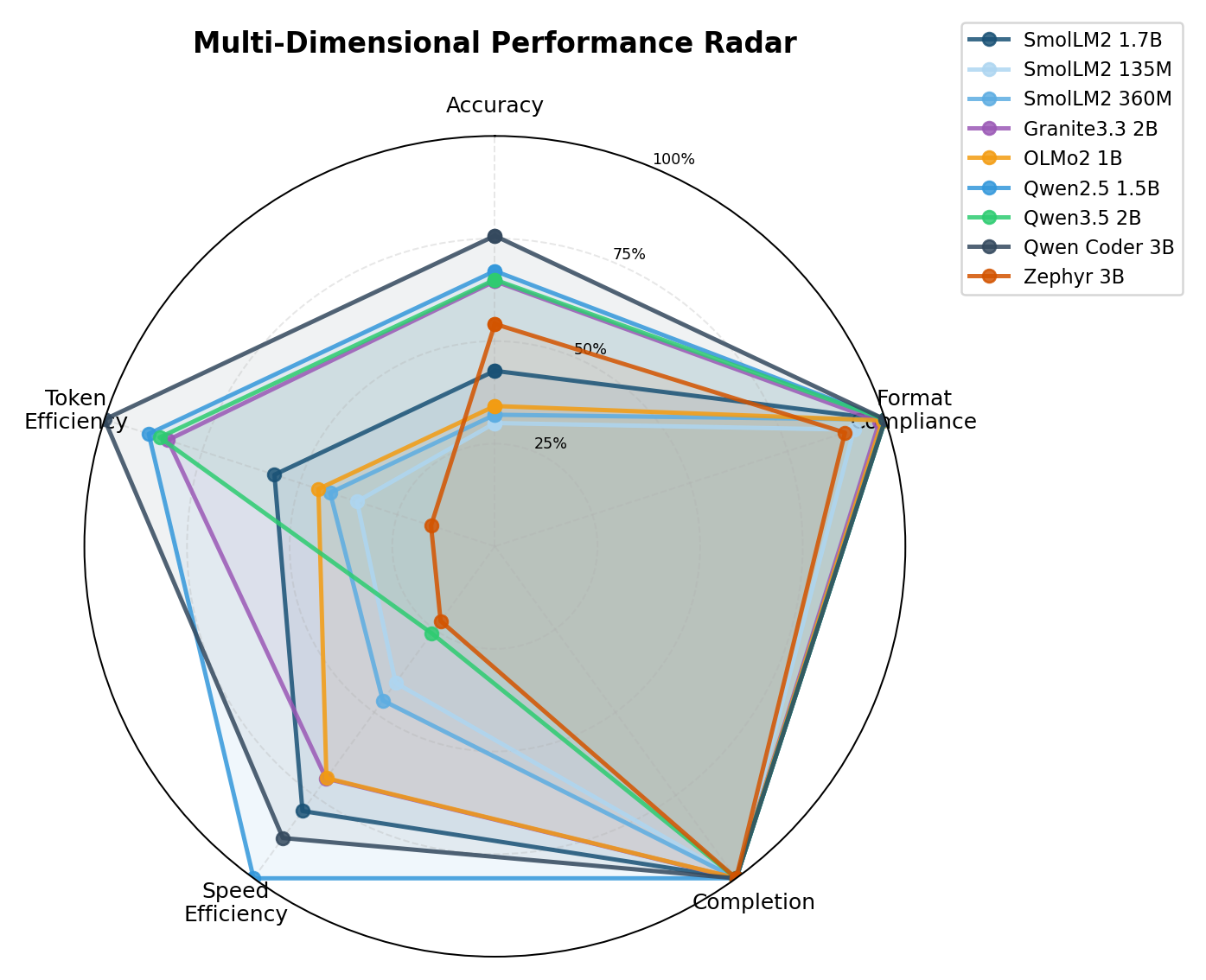}
\caption{Radar chart over accuracy, format compliance, completion, speed efficiency, and token efficiency. Similar global accuracies can conceal very different operational profiles.}
\label{fig:radar}
\end{figure}

The multi-objective view in Figure~\ref{fig:radar} shows why accuracy alone is an incomplete deployment criterion. The strongest models in strict accuracy do not necessarily dominate token efficiency or latency efficiency, and some lower-ranked models exhibit attractive operational geometry despite weaker semantic scores. For local deployment, these profile differences matter because the bottleneck may be latency, memory, cost per request, or aggregate throughput rather than absolute accuracy.

\begin{figure*}[!tbp]
\centering
\includegraphics[width=0.97\textwidth]{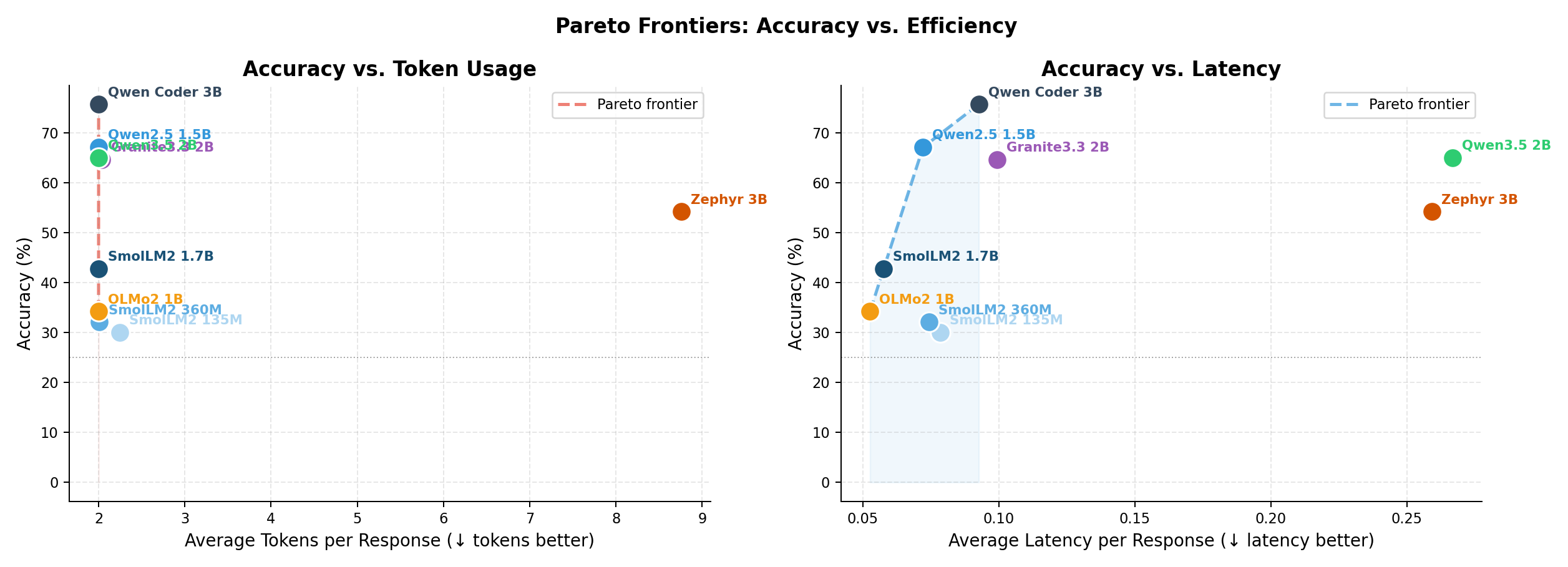}
\caption{Pareto frontiers for accuracy versus token cost and latency. The frontier isolates deployment candidates that remain non-dominated under resource-aware selection rules.}
\label{fig:pareto}
\end{figure*}

Pareto analysis in Figure~\ref{fig:pareto} makes the deployment trade-off explicit. A model on the frontier is non-dominated under a given objective pair; moving away from the frontier means accepting strictly worse accuracy-cost trade-offs. The token and latency panels are not identical, which is itself informative: a token-efficient model need not be the best latency choice, and vice versa.

\begin{figure*}[!tbp]
\centering
\includegraphics[width=0.97\textwidth]{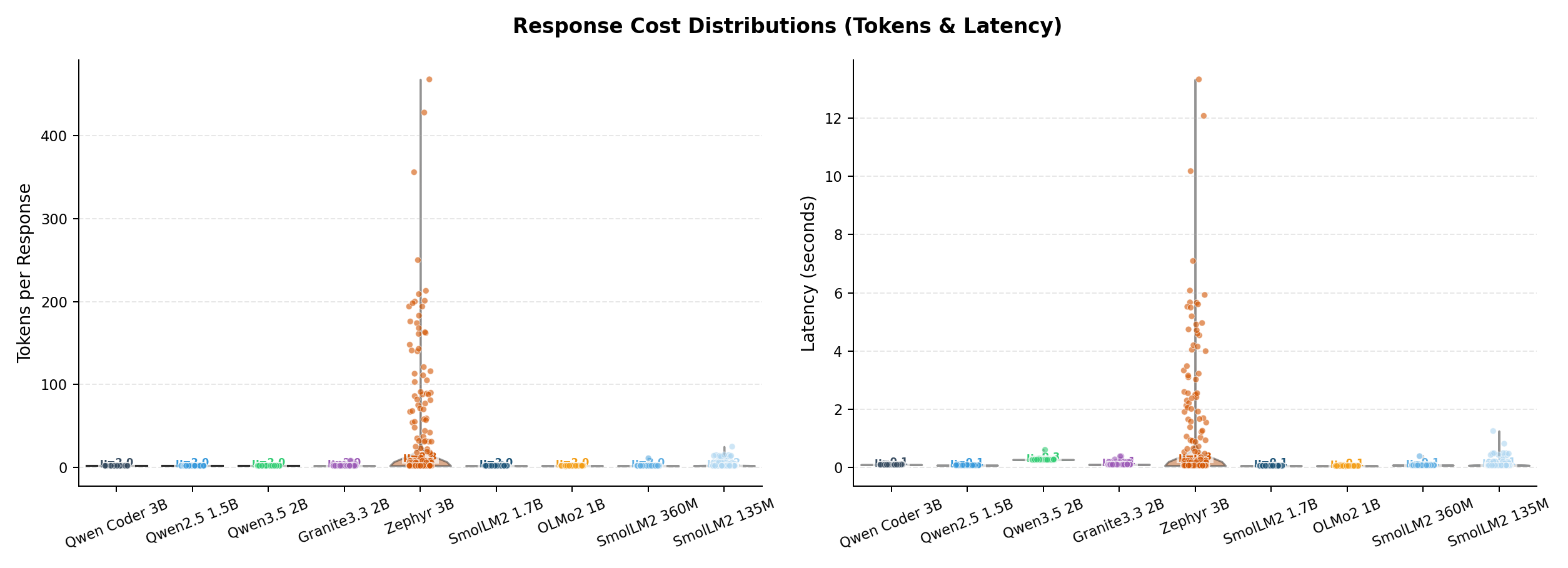}
\caption{Token and latency distributions in linear and log-scale views. Tail behavior is materially different across models and affects deployment desirability beyond mean efficiency.}
\label{fig:costs}
\end{figure*}

Figure~\ref{fig:costs} complements this with full response-cost distributions in both linear and log views. This matters because deployment risk is often governed by tails rather than means: p95 and p99 latency are more consequential than average latency for interactive systems, and heavy-tailed token behavior can dominate cost even when the median is small.

Taken together, the retained base-evaluation views support a stronger claim than a simple leaderboard would. A subset of compact models is already viable for structured workloads, but viability is intrinsically multi-objective and task-conditional. That is exactly the setting in which local experts are preferable to one-size-fits-all model selection.

\FloatBarrier

\section{Fine-Tuning Results}
\subsection{Model-Level Gains on the Shared Held-Out Split}
Fine-tuning is evaluated on a shared 108-example held-out partition produced by the same topic-stratified split for every adapted model. Because the test set is smaller than the full benchmark, Table~\ref{tab:ft} reports Wilson 95\% confidence intervals for both base and post-tuning accuracy.

\begin{table}[t]
\centering
\small
\resizebox{\linewidth}{!}{
\begin{tabular}{lrrrrr}
\toprule
Model & Base (\%) & Base 95\% CI & FT (\%) & FT 95\% CI & Delta (pts) \\
\midrule
Qwen Coder 3B & 65.74 & [56.4, 74.0] & 92.59 & [86.1, 96.2] & +26.85 \\
SmolLM2 1.7B & 46.30 & [37.2, 55.7] & 72.22 & [63.1, 79.8] & +25.92 \\
Qwen2.5 1.5B & 70.37 & [61.2, 78.2] & 89.81 & [82.7, 94.2] & +19.44 \\
SmolLM2 360M & 35.19 & [26.8, 44.6] & 45.37 & [36.3, 54.8] & +10.18 \\
SmolLM2 135M & 25.93 & [18.6, 34.9] & 31.48 & [23.5, 40.7] & +5.55 \\
\bottomrule
\end{tabular}
}
\caption{Base-to-fine-tuned deltas for models with positive observed gains on the shared 108-example test partition, with Wilson 95\% confidence intervals.}
\label{tab:ft}
\end{table}

\begin{figure*}[!tbp]
\centering
\includegraphics[width=0.97\textwidth]{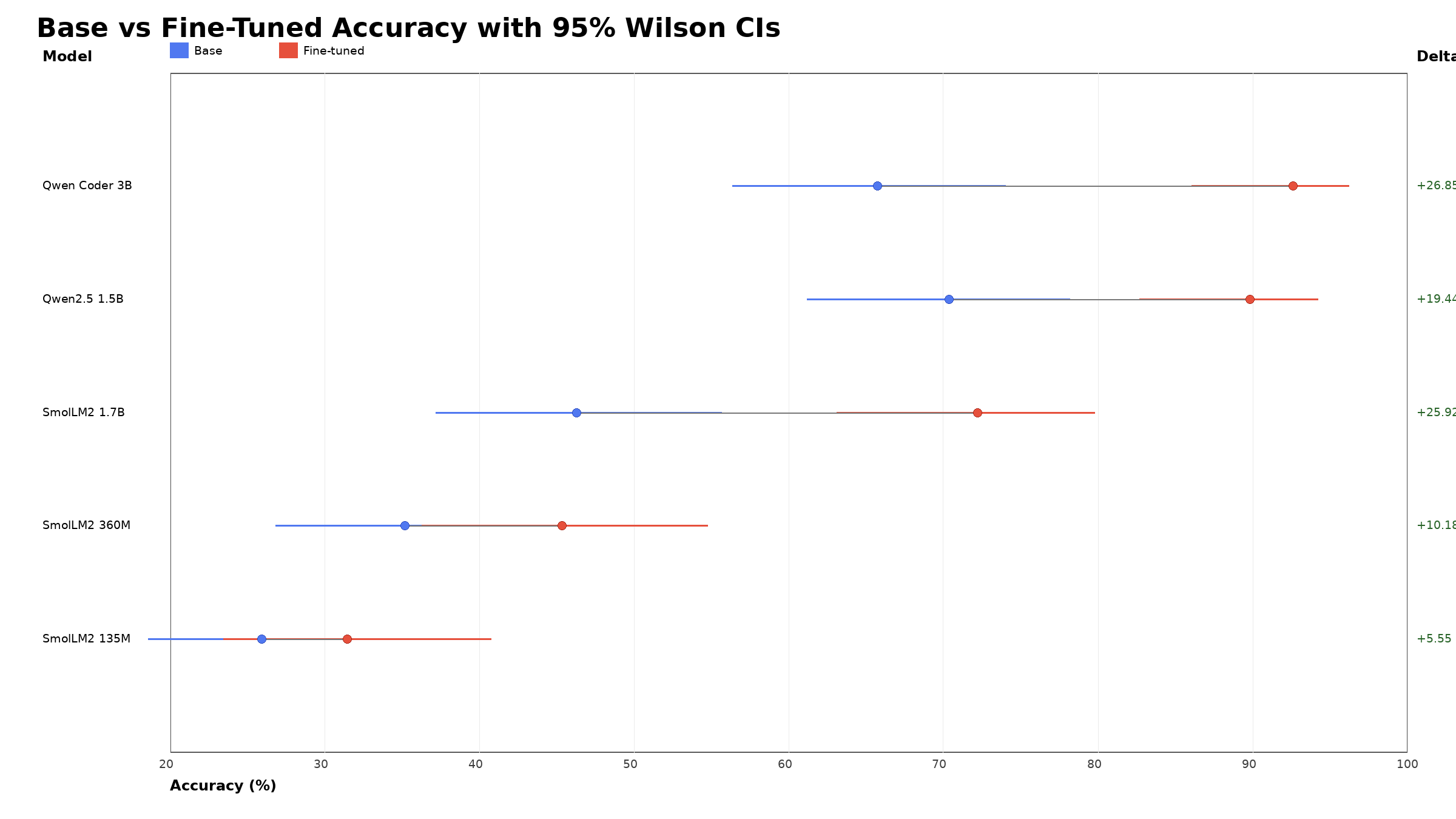}
\caption{Base versus fine-tuned strict accuracy with 95\% Wilson confidence intervals on the shared held-out split. The gray connectors visualize per-model gain directly.}
\label{fig:ft_dumbbell}
\end{figure*}

Figure~\ref{fig:ft_dumbbell} makes the same pattern visually explicit. The largest improvements are concentrated in Qwen Coder 3B, SmolLM2 1.7B, and Qwen2.5 1.5B. The qualitative interpretation is important: the strongest gains occur when the base model already has excellent format control but still exhibits semantic headroom. PEFT is therefore acting primarily as decision-boundary specialization rather than as a general repair mechanism.

This interpretation is also consistent with the absolute ceilings. The smallest SmolLM2 models improve, but remain bounded by representational capacity. Qwen Coder 3B, by contrast, combines a strong base representation with sufficient adapter leverage to move into a clearly higher operating regime. For deployment, that distinction matters more than raw delta alone: a large gain is only practically valuable if the final accuracy clears the use-case threshold.

\subsection{Topic-Level Transfer and Aggregate Effect}
\begin{figure*}[!tbp]
\centering
\includegraphics[width=0.97\textwidth]{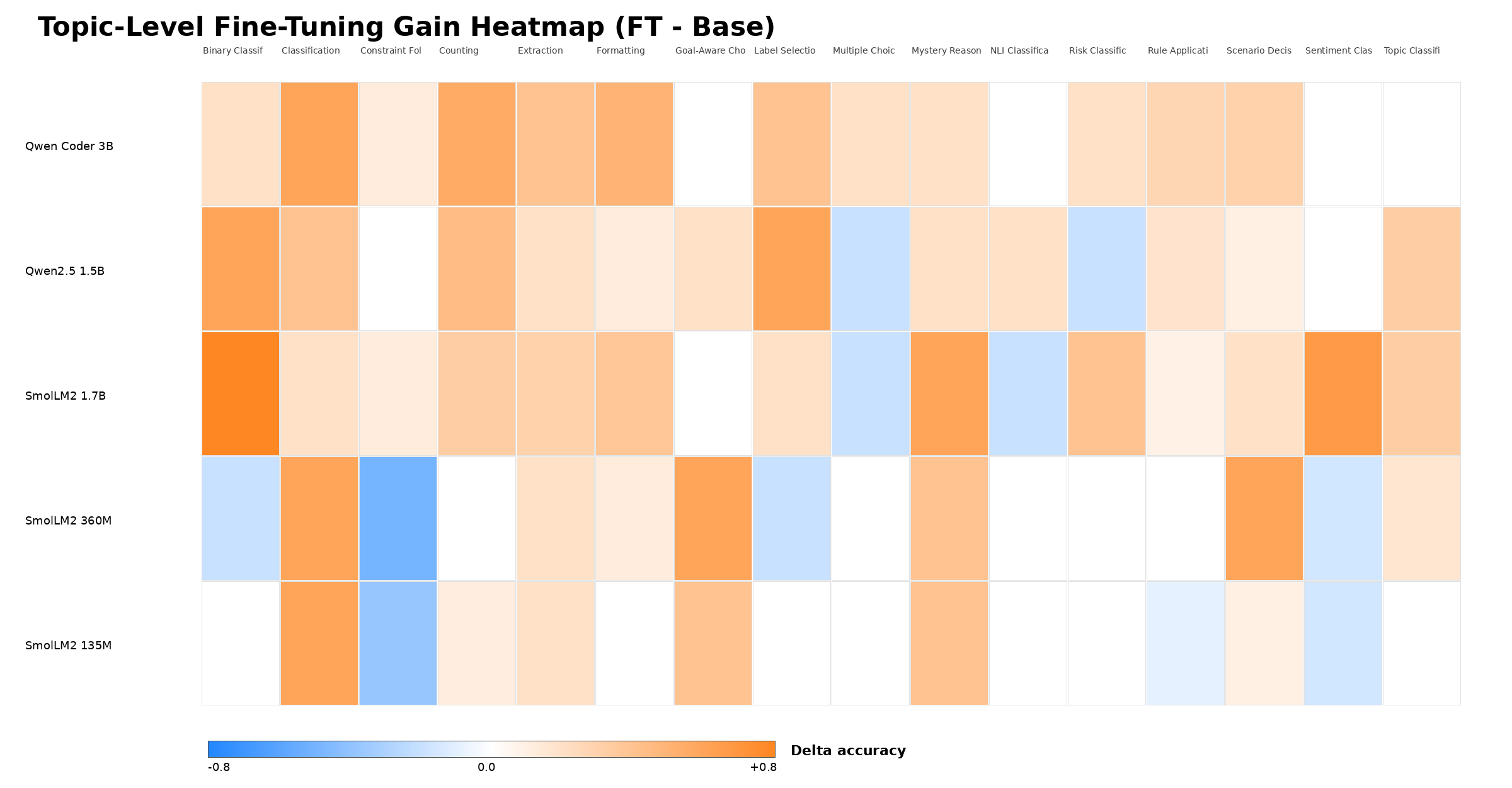}
\caption{Topic-level fine-tuning deltas by model. Warm colors indicate positive transfer and cool colors indicate regressions, making the heterogeneity of adaptation visible at task-family resolution.}
\label{fig:ft_heatmap}
\end{figure*}

Aggregate uplift is substantial: across the five models with positive observed gains, the mean improvement is +17.59 points. But the more informative question is whether transfer is uniform. Figure~\ref{fig:ft_heatmap} shows that it is not. Positive transfer is broad, especially for Qwen Coder 3B and SmolLM2 1.7B, but the heatmap also contains pockets of flat or negative movement. Fine-tuning therefore redistributes competence across the topic matrix rather than monotonically improving every cell.

The strongest average topic gains occur in \textit{Classification}, \textit{Mystery Reasoning}, \textit{Counting}, \textit{Extraction}, and \textit{Scenario Decision}. Weaker or negative shifts in \textit{Constraint Following} and \textit{Multiple Choice Reasoning} indicate that the adapter updates are not purely additive; they reshape a shared representational space. Under a constrained rank budget, sharpening one local manifold can slightly degrade another.

At model granularity, Qwen Coder 3B shows the most convincing combination of breadth and final accuracy, suggesting that its pretrained representation is especially amenable to task-specific refinement under completion-only supervision. SmolLM2 1.7B exhibits the clearest ``headroom + adaptability'' profile: the base score leaves substantial room for improvement, yet the representation is strong enough to capitalize on the adapter updates. Qwen2.5 1.5B is perhaps the most practically attractive compromise for many local deployments because it begins from a strong base and still admits a large positive shift.

\begin{figure*}[!tbp]
\centering
\includegraphics[width=0.97\textwidth]{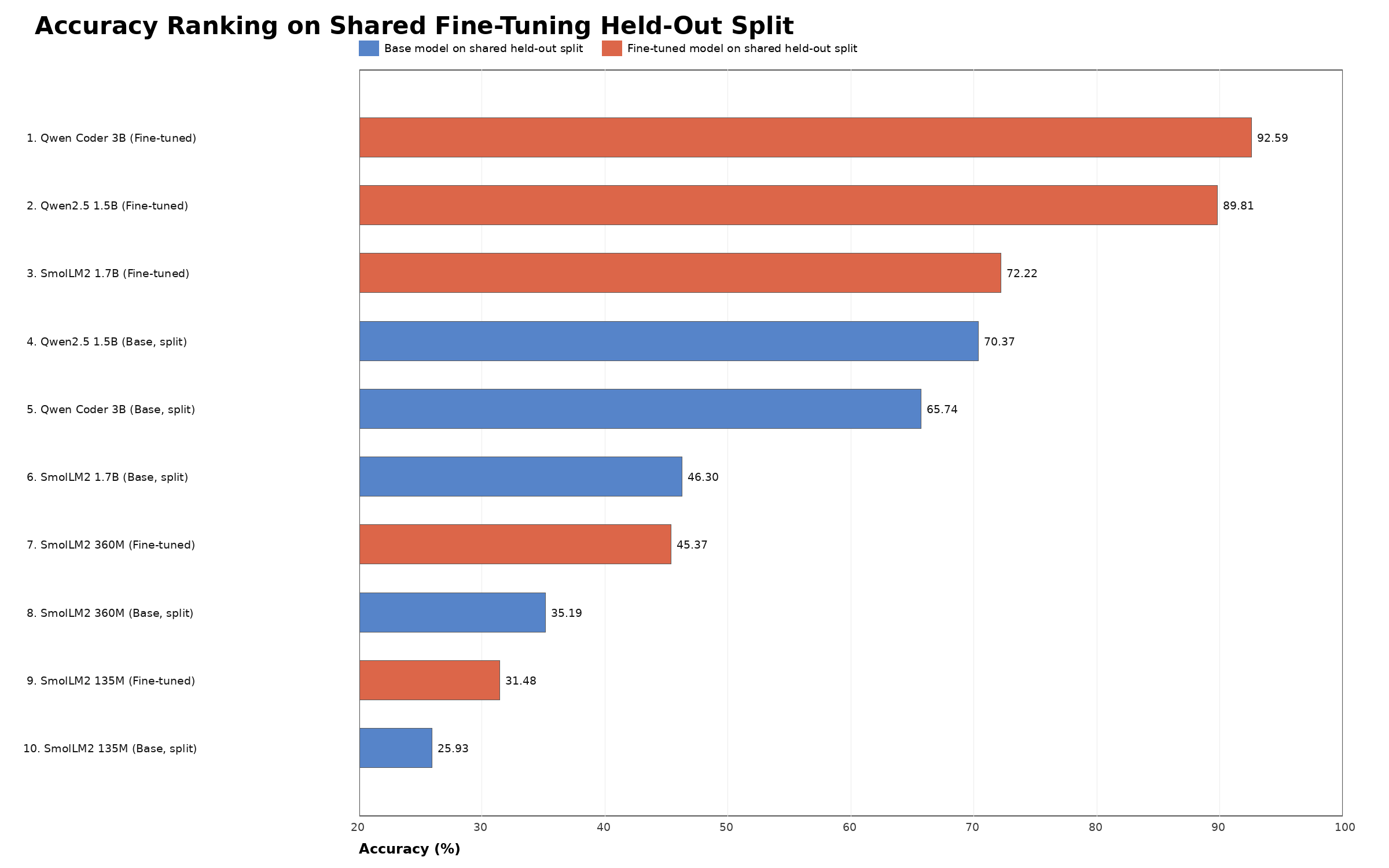}
\caption{Split-consistent ranking of base and fine-tuned variants on the shared fine-tuning held-out split. Only the five models with completed fine-tuning runs are included.}
\label{fig:ft_ranking}
\end{figure*}

To keep the ranking view split-consistent, Figure~\ref{fig:ft_ranking} compares base and fine-tuned variants only on the same held-out partition used in Table~\ref{tab:ft}. This avoids mixing full-benchmark base scores with split-level post-tuning scores. Under this matched view, Qwen2.5 1.5B begins slightly above Qwen Coder 3B before being overtaken after fine-tuning, which is why the adapted ranking is not identical to the full 1,085-example base leaderboard.

The broader lesson is methodological. PEFT should be treated neither as a universal guarantee nor as a marginal optimization. On this benchmark it is frequently decisive, but only when evaluated under the same split, scoring rule, and deployment contract as the base model. That is why the correct unit of analysis is not ``Did fine-tuning help?'' in the abstract, but ``For which model-task families does low-cost specialization materially improve viability?''

\FloatBarrier

\section{Discussion}
\subsection{Local Experts as a Democratization Primitive}
The evidence supports a specific account of AI democratization. Democratization does not require that small models rival the best frontier systems on every task. It requires that useful competence be attainable through an auditable, low-cost process that smaller organizations can own. The results here show that this is already plausible for structured workloads. Several base models are serviceable before adaptation, and adapter-based specialization can convert some of them into strong local experts.

That finding has both systems and governance implications. From a systems perspective, local deployment can be organized around a portfolio rather than a monolith: choose compact models whose failure surfaces and efficiency profiles match the intended workload, adapt them with low-rank updates, and route only the relevant task families to each checkpoint. From a governance perspective, this architecture increases data control, reduces dependency on external API providers, and gives smaller institutions a practical path to experimentation and ownership.

The study therefore argues for a workflow, not a winner-take-all leaderboard. The relevant pipeline is: identify a bounded niche task, curate an evaluation set that reflects that task, run consistent cross-model evaluation, select candidates under multi-objective constraints, fine-tune the most promising models, and then validate topic by topic before deployment. The benchmark introduced here can function as a reusable baseline, but the more general contribution is the procedure itself.

\subsection{Scope, Limitations, and Boundary Conditions}
The empirical claims are intentionally benchmark-conditional. The reported rankings and fine-tuning deltas apply to this dataset, this prompt scaffold, this deterministic decoding setup, and this strict output-validity rule. The paper does \emph{not} claim universal superiority for any specific model family beyond that scope.

Several limitations follow directly. First, the fine-tuning held-out split contains 108 examples, so uncertainty is non-trivial even with Wilson intervals. Second, the benchmark mixes binary, ternary, and four-way choice spaces, which complicates any attempt to reduce performance to a chance-adjusted scalar. Third, the benchmark is template-structured by design: it improves comparability and reproducibility, but it underrepresents the linguistic diversity of truly open-ended user traffic.

These are not incidental caveats; they define the inferential boundary of the study. The appropriate interpretation is that the paper establishes a high-signal baseline for structured, template-aligned MCQ settings and demonstrates a replicable specialization workflow. Extending the same workflow to broader domains requires new datasets, new failure analyses, and potentially new adaptation recipes.

\section{Reproducibility Statement}
The study is intentionally reproducibility-oriented. The benchmark is explicit at the row level: each item contains a topic label, question, option structure, full prompt scaffold, and gold answer. Benchmark construction follows a category-first generator process with fixed target counts, controlled instruction templates, and schema checks. In v3, the released specification is 1,085 rows across 16 topics, 158 unique question stems, 611 A/B/C/D items, 312 A/B items, and 162 A/B/C items, with final label frequencies A=413, B=353, C=200, and D=119.

Evaluation is deterministic at the decoding and scoring layers. Fine-tuning uses a shared split seed, a fixed PEFT configuration, and the same strict correctness rule before and after adaptation. The figures in this manuscript are generated from saved evaluation artifacts rather than from ad hoc manual summaries. The benchmark can therefore be reused directly as a baseline, or extended with new topic families and harder distractor structures under the same schema.

\section{Broader Impact Statement}
The same properties that make compact local models attractive also lower barriers to misuse. Lightweight local deployment can enable harmful automation, and structured outputs can create over-trust if apparent format compliance is mistaken for deep competence. Topic-specific brittleness, class imbalance, and dataset-conditioned overfitting are all risks if the models are deployed without calibration or post-deployment auditing.

The positive case is equally real. Local experts can keep sensitive data on controlled infrastructure, reduce recurring API costs, support institution-specific customization, and broaden participation for educational, civic, and low-resource settings that cannot support frontier-model serving. The practical upside of democratization is therefore not only lower cost, but broader agency over model selection, adaptation, and governance.

Responsible use in high-stakes settings still requires safeguards: task scoping, output validation, abstention policies, and periodic audits for bias concentration and drift. Compactness is not a substitute for safety discipline.

\section{Future Work}
Three directions are immediate. The first is statistical strengthening: repeated split evaluation, multi-seed fine-tuning, and larger held-out sets would tighten uncertainty around PEFT gains. The second is robustness: option-order shuffling, prompt perturbation, calibration analysis, and abstention-aware decoding would reveal whether the measured competence is stable under interface variation. The third is benchmark expansion: harder distractors, longer context windows, and more linguistically diverse stems would better separate shallow template adaptation from durable reasoning competence.

A broader systems direction is equally important. The local-expert hypothesis should be tested on real organizational workloads where privacy, governance, and total cost of ownership can be measured directly rather than inferred from proxy metrics.

\section{Conclusion}
The results support a clear but bounded conclusion. A subset of compact open models is already viable for structured local deployment, and parameter-efficient specialization can materially expand that viability without requiring frontier-scale infrastructure. The relevant contribution is not the discovery of a single universally best SLM. It is the demonstration that disciplined evaluation plus low-cost adaptation provides a practical route to community-controlled AI systems. In that sense, democratization is not a rhetorical add-on to compact models; it is the operational consequence of being able to run, audit, specialize, and govern them locally.

\section*{Acknowledgments}
Acknowledgment is given to the open-model and open-tooling communities for enabling reproducible SLM research, including maintainers of model cards, PEFT stacks, and evaluation infrastructure used throughout this study.

\section*{References}
\begin{enumerate}[leftmargin=*]
\item E. J. Hu et al. ``LoRA: Low-Rank Adaptation of Large Language Models.'' ICLR, 2022.
\item T. Dettmers et al. ``QLoRA: Efficient Finetuning of Quantized LLMs.'' arXiv:2305.14314, 2023.
\item S.-Y. Liu et al. ``DoRA: Weight-Decomposed Low-Rank Adaptation.'' arXiv:2402.09353, 2024.
\item L. B. Allal et al. ``SmolLM2: When Smol Goes Big -- Data-Centric Training of a Small Language Model.'' arXiv:2502.02737, 2025.
\item L. Soldaini et al. ``2 OLMo 2 Furious.'' arXiv:2501.00656, 2025.
\item A. Yang et al. ``Qwen2 Technical Report.'' arXiv:2407.10671, 2024.
\item Qwen Team. ``Qwen2.5-Coder Technical Report.'' arXiv:2409.12186, 2024.
\item Hugging FaceTB. ``SmolLM2-135M-Instruct Model Card.'' 2025.
\item Hugging FaceTB. ``SmolLM2-360M-Instruct Model Card.'' 2025.
\item Hugging FaceTB. ``SmolLM2-1.7B-Instruct Model Card.'' 2025.
\item Stability AI. ``stablelm-zephyr-3b Model Card.'' 2024.
\item Qwen Team. ``Qwen2.5-1.5B-Instruct Model Card.'' 2024.
\item Qwen Team. ``Qwen2.5-Coder-3B-Instruct Model Card.'' 2024.
\item Qwen Team. ``Qwen3.5-2B Model Card.'' 2026.
\item IBM Granite Team. ``granite-3.3-2b-instruct Model Card.'' 2025.
\item Allen Institute for AI. ``OLMo-2-0425-1B-Instruct Model Card.'' 2025.
\item Hugging Face. ``Transformers Quantization with bitsandbytes Documentation.'' 2026.
\item Hugging Face. ``PEFT LoRA Developer Guide.'' 2026.
\item Hugging Face. ``TRL SFT Trainer Documentation.'' 2026.
\item NVIDIA. ``NVIDIA L4 Tensor Core GPU Specifications.'' 2026.
\item R. Bommasani et al. ``On the Opportunities and Risks of Foundation Models.'' arXiv:2108.07258, 2021.
\item E. M. Bender et al. ``On the Dangers of Stochastic Parrots.'' FAccT, 2021.
\item E. Strubell, A. Ganesh, and A. McCallum. ``Energy and Policy Considerations for Deep Learning in NLP.'' ACL, 2019.
\item D. Patterson et al. ``Carbon Emissions and Large Neural Network Training.'' arXiv:2104.10350, 2021.
\item L. Chen, M. Zaharia, and J. Zou. ``FrugalGPT: How to Use Large Language Models While Reducing Cost and Improving Performance.'' arXiv:2305.05176, 2023.
\item P. Zhang et al. ``TinyLlama: An Open-Source Small Language Model.'' arXiv:2401.02385, 2024.
\item M. Abdin et al. ``Phi-3 Technical Report: A Highly Capable Language Model Locally on Your Phone.'' arXiv:2404.14219, 2024.
\item Z. Liu et al. ``MobileLLM: Optimizing Sub-Billion Parameter Language Models for On-Device Use Cases.'' arXiv:2402.14905, 2024.
\end{enumerate}

\end{document}